\documentclass[conference]{IEEEtran}
\usepackage{times}

\usepackage[numbers]{natbib}
\usepackage{multicol}
\usepackage[bookmarks=true]{hyperref}

\usepackage{multirow}
\usepackage{graphicx} 

\usepackage{caption}
\usepackage{subcaption}

\usepackage{xcolor}
\usepackage{algorithm}
\usepackage[noend]{algpseudocode} 
\usepackage{amssymb,amsmath,comment,cite}

\newtheorem{definition}{Definition}
\usepackage{url}
\usepackage{comment}

\usepackage{ifthen}
\newboolean{shortver}
\setboolean{shortver}{false}

\begin{document}



\title{A Local Optimization Framework for Multi-Objective Ergodic Search}





\author{\authorblockN{Zhongqiang Ren\authorrefmark{1},
Akshaya Kesarimangalam Srinivasan\authorrefmark{1},
Howard Coffin\authorrefmark{1}, 
Ian Abraham\authorrefmark{2} and
Howie Choset\authorrefmark{1}}
\authorblockA{\authorrefmark{1}Carnegie Mellon University, 5000 Forbes Ave., Pittsburgh, PA 15213.\\ Email: \{zhongqir, akesarim, hcoffin, choset\}@andrew.cmu.edu}
\authorblockA{\authorrefmark{2}Yale University, 17 Hillhouse Avenue, New Haven, CT 06511.
Email: ian.abraham@yale.edu}}



%

\newcommand{\red}{\color{red}}
\newcommand{\green}{\color{green}}
\newcommand{\blue}{\color{blue}}

\maketitle

\thispagestyle{plain}
\pagestyle{plain}
\pagenumbering{arabic}

\begin{abstract}
Robots have the potential to perform search for a variety of applications under different scenarios. 
Our work is motivated by humanitarian assistant and disaster relief (HADR) where often it is critical to find signs of life in the presence of conflicting criteria, objectives, and information. 
We believe ergodic search can provide a framework for exploiting available information as well as exploring for new information for applications such as HADR, especially when time is of the essence. 
Ergodic search algorithms plan trajectories such that the time spent in a region is proportional to the amount of information in that region, and is able to naturally balance exploitation (myopically searching high-information areas) and exploration (visiting all locations in the search space for new information).
Existing ergodic search algorithms, as well as other information-based approaches, typically consider search using only a single information map.
However, in many scenarios, the use of multiple information maps that encode different types of relevant information is common. 
Ergodic search methods currently do not possess the ability for simultaneous nor do they have a way to balance which information gets priority.
This leads us to formulate a Multi-Objective Ergodic Search (MOES) problem, which aims at finding the so-called Pareto-optimal solutions, for the purpose of providing human decision makers various solutions that trade off between conflicting criteria.
To efficiently solve MOES, we develop a framework called Sequential Local Ergodic Search (SLES) that converts a MOES problem into a ``weight space coverage'' problem. It leverages the recent advances in ergodic search methods as well as the idea of local optimization to efficiently approximate the Pareto-optimal front.
Our numerical results show that SLES computes solutions of better quality than the popular multi-objective genetic algorithms and runs distinctly faster than a naive scalarization method on a commercial laptop.

\end{abstract}


\section{Introduction}\label{moes:sec:intro}
\graphicspath{{figures/}}

\begin{figure}[t!]
	\centering
	\includegraphics[width=\linewidth]{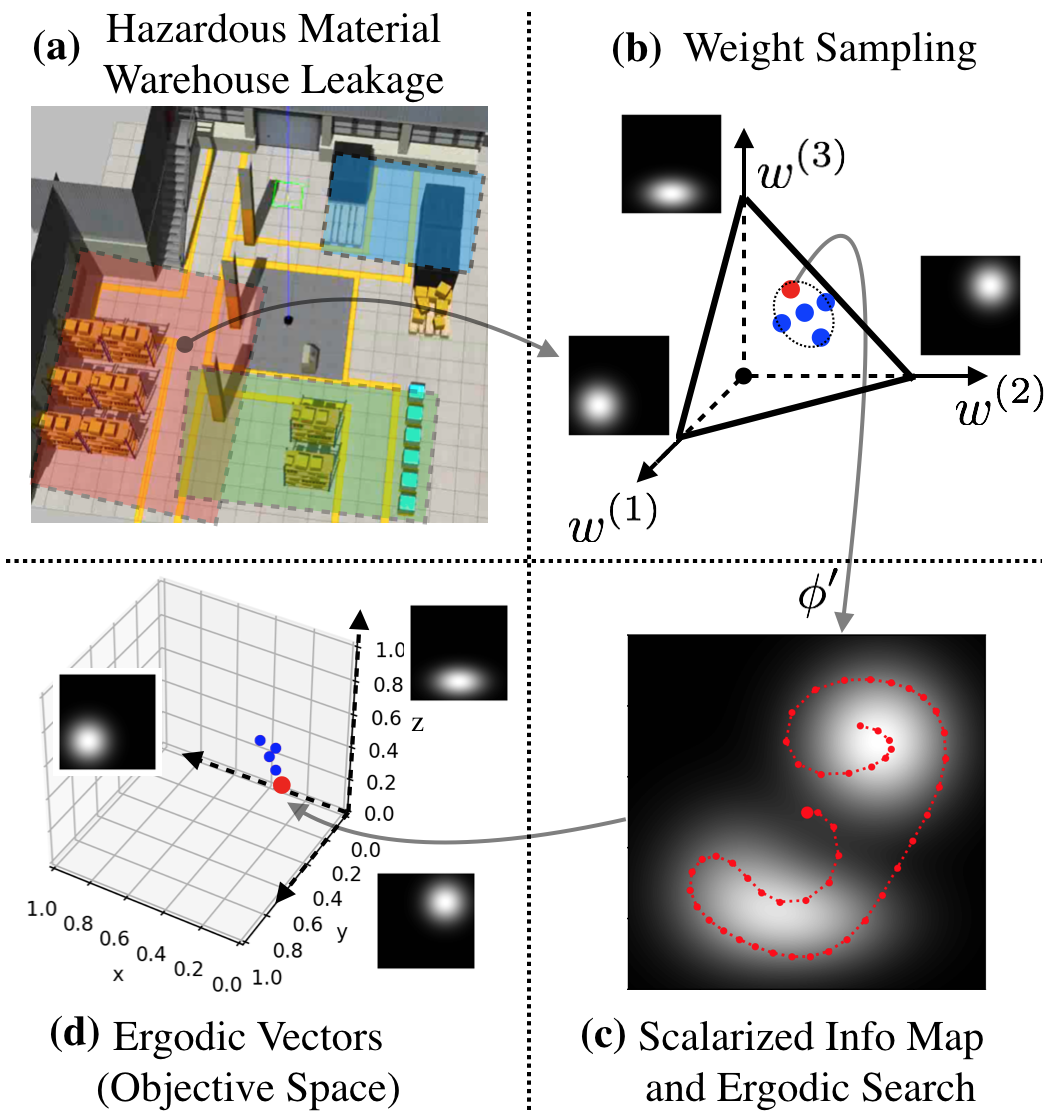}
	\caption{A visualization of the MOES problem and our method. (a) shows a search and rescue task in a hazardous material warehouse with leakage, where colored areas indicate different types of information/targets such as survivors, leakage sources, etc. Each type of information is represented as an info map. (b) shows the weight space $\mathcal{B}$ in the presence of three objectives, where $w^{(i)}$ is the relative weight of the corresponding info map $\phi^{(i)}$, with $i=1,2,3$. (c) shows the scalarized info map $\phi'$, which is the weighted-sum of all three info maps. An ergodic trajectory is planned with respect to $\phi'$. (d) shows the objective space, where each element is an ergodic vector that describes the ergodic metric of the computed trajectory with respect to $\phi^{(1)},\phi^{(2)},\phi^{(3)}$. The computed ergodic vectors approximate the Pareto-optimal front.}
	\label{moes:fig:algo_overview}
	\vspace{-3mm}
\end{figure}

This work considers the motion planning problem for area search/coverage, which arises in many applications such as search and rescue~\citep{lee2018receding,liu2013robotic}, environment monitoring~\citep{Liu-RSS-19}, target localization~\citep{garzon2016multirobot,pimenta2009simultaneous}.
Given an information map (abbreviated as info map), which describes the prior knowledge\footnote{If no prior knowledge is available, a uniform information distribution can be used.} about the information distribution over the area to be covered, the problem requires planning a trajectory to efficiently gather information.
Common approaches to this problem span a spectrum from spatial decomposition methods~\citep{acar2002morse,santos2018coverage,schwager2009decentralized}, which uniformly cover the area, to information-theoretic approaches~\citep{Liu-RSS-19,julian2012distributed}, which greedily move the robot to the next location with the highest information gain.
This work is interested in ergodic search algorithms~\citep{mathew2011metrics,miller2013trajectory,miller2016ergodic,salman2017multi}, which lie in the middle of the spectrum:
this type of methods optimizes an ergodic metric to plan trajectories along which the time spent in a region is proportional to the amount of information in that region.
Ergodic search inherently balances exploitation and exploration.

Existing ergodic search algorithms consider covering only a single info map.
However, in many applications, multiple different info maps, each of which encodes one type of information, may need to be covered simultaneously.
As an example, consider a search and rescue task in a hazardous material warehouse with leakage where a robot is deployed to search for both survivors and leakage sources (Fig.~\ref{moes:fig:algo_overview} (a)).
Multiple info maps describing probable locations of survivors and leakage sources are required to be simultaneously covered.
Additionally, these info maps may not be readily \emph{scalarized} and added into a single info map (via weighted-sum for example), since the relative importance between them is unknown or hard to obtain.
In this work, we formulate a Multi-Objective Ergodic Search (MOES) problem to describe such scenarios, which requires planning trajectories that can simultaneously cover multiple info maps.
In general, there is no single trajectory that optimizes the ergodic metrics with regard to all info maps at the same time.
Thus, this work seeks to find a set of Pareto-optimal solution (trajectories): a solution is Pareto-optimal if one can not improve the ergodic metric with respect to one info map without deteriorating the ergodic metric with regard to at least one of the other info maps.
We believe the visualization of a set of Pareto-optimal solutions can help the human decision makers (who are often involved in search and rescue tasks~\citep{liu2013robotic}) make more informed decisions based on their domain knowledge.

Baseline approaches that can be used to solve MOES include general-purpose multi-objective genetic algorithms (MOGA)~\citep{deb2002fast,emmerich2018tutorial,zhang2007moea}.
While being applicable to various problems, MOGAs typically fail to leverage the underlying structure of MOES problems (such as the dynamics of the robot and local metric structures e.g., ``convexity''), which can make them inefficient to optimize.
Another baseline is the scalarization method~\citep{ehrgott2005multicriteria,emmerich2018tutorial}, which can be applied to solve MOES by sampling a set of weight vectors, computing the weighted-sum of all the info maps (referred to as a scalarized info map) for each weight vector, and running regular (single-objective) ergodic search for each of the scalarized info map in an episodic manner.
While being able to exploit the underlying problem structure, the scalarization method can be time-consuming due to the episodic computation corresponding to weight vectors, especially when many different Pareto-optimal solutions are desired.

In this work, we take the view that an approximated Pareto-optimal set of solutions can be efficiently obtained by leveraging \emph{local optimization} based on the inherent convexity of the ergodic metric in the Fourier coefficient space, and develop a framework called Sequential Local Ergodic Search (SLES), which is conceptually visualized in Fig.~\ref{moes:fig:algo_overview}.
SLES resembles scalarization methods in episodically computing weighted-sum of the info maps.
However, SLES ``covers'' the weight space (the space that contains all possible weight vectors) in a breadth-first manner by (i) sampling new weight vectors in the neighborhood of the current weight vector, and (ii) optimizing the trajectory corresponding to the new weight vector by using the current solution as an initial guess (to warm-start the optimization).
To expedite the coverage of the weight space by SLES without sacrificing the solution quality, we also develop a variant called Adaptive SLES (A-SLES), which can adjust the density of sampled weight vectors based on the ``similarity'' of info maps to be covered. 
Our numerical results show that SLES and A-SLES compute a set of solution trajectories with better ergodic metrics in comparison with naively applying MOGAs to MOES. Additionally, SLES and A-SLES require less than half of the run time of a naive scalarization method without sacrificing the quality of the solutions.
We also simulate our method for a search and rescue task in a hazardous material warehouse in ROS/Gazebo.

The rest of this article is organized as follows.
Sec.~\ref{moes:sec:related} briefly reviews related work and Sec.~\ref{moes:sec:preli} introduces basic concepts and the problem definition.
We then elaborate our idea and method in Sec.~\ref{moes:sec:method}, discuss the numerical results in Sec.~\ref{moes:sec:result}, and conclude in Sec.~\ref{moes:sec:conclusion}.

\section{Related Work}\label{moes:sec:related}

\noindent\textbf{Ergodic Coverage}.
A trajectory is ergodic with respect to an info map if the amount of time spent in a region is proportional to the amount of information in that region.
Ergodic metrics, such as~\citep{mathew2011metrics}, measure how far a trajectory is from being ergodic, and by iteratively minimizing the metric, an ergodic trajectory can be computed~\citep{mathew2011metrics}.
Ergodic trajectory planning has been investigated within the framework of receding horizon control~\citep{miller2013trajectory}, stochastic optimization~\citep{ayvali2017ergodic}, and has been leveraged for active learning and search~\citep{abraham2021ergodic,miller2015ergodic}, decentralized exploration~\citep{abraham2018decentralized},  real-time area coverage and target localization~\citep{mavrommati2017real}, etc.
However, we are not aware of any ergodic search method that considers covering multiple info maps at the same time, which is the focus of this work.

\vspace{2mm}
\noindent\textbf{Multi-Objective Optimization} (MOO) is a broad topic~\citep{ehrgott2005multicriteria,emmerich2018tutorial} and has been investigated in robotics-related problems such as path planning~\citep{ren21mocbs,ren21momstar,ren22mopbd}, reinforcement learning~\citep{roijers2013survey}, and design~\citep{nardi2019practical}.
With respect to MOO for coverage/search tasks, existing work has considered simultaneously optimizing exploration and exploitation for environment monitoring tasks~\citep{Liu-RSS-19}.
\citet{lee2018receding} optimizes the coverage of a single info map using the ergodic metric while optimizing other ``non-ergodic'' objectives by using $\epsilon$-constraints.
Our work differs from them, as we aim to ergodically cover multiple info maps, each of which represents an objective.
As shown in Sec.~\ref{moes:sec:method:basic}, this problem formulation allows us to leverage the inherent convexity of the ergodic metric in the Fourier coefficient space to efficiently obtain an approximated Pareto-optimal front.

\section{Preliminaries}\label{moes:sec:preli}
\graphicspath{{figures/}}

\subsection{Ergodic Metric}

Let $\mathcal{W}=[0,L_1]\times[0,L_2]\times\dots\times[0,L_\nu] \subset \mathbb{R}^\nu$ denote a $\nu$-dimensional workspace that is to be explored by a robot.
The robot has a $n$-dimensional state space ($n\geq\nu$), and let $q_n:[0,T]\rightarrow\mathbb{R}^n$ denote a trajectory in the state space with $T\in\mathbb{R}^+$ representing the time horizon.
The robot (deterministic) dynamics is described as $\dot{q_n}(t)=f(q_n(t),u(t))$, where $u(t)$ is the control input of the robot.
Additionally, for each trajectory $q_n$, let $q:[0,T]\rightarrow\mathcal{W}$ denote the corresponding trajectory in the workspace (instead of the state space).

Let $c(x,q), x\in\mathcal{W}$ denote the time-averaged statistics of a trajectory $q$, which is defined as:
\begin{gather}\label{moes:eqn:time_avg_stat_of_traj}
    c(x,q) = \frac{1}{T}\int_{0}^{T} \delta(x-q(\tau))d\tau,
\end{gather} where $\delta$ is a Dirac function.
Let $\phi:\mathcal{W}\rightarrow\mathbb{R}$ denote a static info map (i.e., a probability distribution), which describes the amount of information at each location in the workspace.
An ergodic metric \citep{mathew2011metrics} between $c(x,q)$ and an info map $\phi$ is defined as:
\begin{equation}
\begin{split}
    \mathcal{E}(\phi,q) & = \sum_{k=0}^{K}\lambda_k(c_k - \phi_k)^2\\
    & =\sum_{k=0}^{K}\lambda_k \left( \frac{1}{T} \int_{0}^{T}F_k(q(\tau))d\tau - \phi_k \right) ^2
\end{split}
\end{equation}
where (i) $\phi_k=\int_{\mathcal{W}}\phi(x)F_k(x)dx$ represents the Fourier coefficients of the info map, with $F_k(q) = \frac{1}{h_k}\Pi_{j=1}^{\nu}\cos(\frac{k_j\pi q_j}{L_j})$ being the cosine basis function for some index $k \in \mathbb{N}^\nu$ and $K$ being the number of Fourier bases considered, (ii) $c_k$ denotes the Fourier coefficient of $c(x,q)$, (iii) $h_k$ denotes the normalization factor as defined in \citep{mathew2011metrics}, and (iv) $\lambda_k=(1+||k||^2)^{-\frac{\nu+1}{2}}$ denotes the weight for each corresponding Fourier coefficient.

\subsection{Ergodic Vector and Pareto-Optimality}

This work aims to plan robot trajectories to simultaneously cover multiple info maps.
We use the superscript in $\phi^{(i)}$ to denote a specific info map, with $i=1,2,\dots,m$ where $m$ is a finite number indicating the total number of info maps to be covered.
Let $\vec{\mathcal{E}}(q)=(\mathcal{E}(\phi^{(1)},q),\mathcal{E}(\phi^{(2)},q),\dots,\mathcal{E}(\phi^{(m)},q))$ denote an \emph{ergodic vector}, which describes the ergodic metrics of the trajectory $q$ with respect to all info maps.
To compare any two trajectories, we compare the ergodic vectors corresponding to them using the dominance relation from the multi-objective optimization literature.
\vspace{1.5mm}
\begin{definition}[Dominance~\citep{ehrgott2005multicriteria}]
	Given two vectors $a$ and $b$ of length $m$, $a$ dominates $b$, notationally $a \succeq b$, if and only if $a(j) \leq b(j)$, $\forall j \in \{1,2,\dots,m\}$ and $a(j) < b(j)$, $\exists j \in \{1,2,\dots,m\}$ .
\end{definition}
\vspace{1.5mm}
If $a$ does not dominate $b$, this non-dominance is denoted as $a \nsucceq b$.
In this work, given two trajectories $q_1,q_2$ (with the same time horizon $[0,T]$), we say $q_1$ dominates $q_2$ (denoted as $q_1\succeq q_2$) if $\vec{\mathcal{E}}(q_1) \succeq \vec{\mathcal{E}}(q_2)$.
Any two trajectories are non-dominated (to each other) if the corresponding ergodic vectors do not dominate each other.
Among all feasible trajectories, the set of all non-dominated trajectories is called the {\it Pareto-optimal} (solution) set, and the set of the corresponding ergodic vectors is called the Pareto-optimal front.

\vspace{1.5mm}
\noindent\textbf{Problem Statement.} This work considers a \emph{Multi-Objective Ergodic Search (MOES)} problem, which requires computing a set of trajectories, whose ergodic vectors approximate the Pareto-optimal front.

\section{Method}\label{moes:sec:method}

\subsection{Basic Concepts and Overview}\label{moes:sec:method:basic}

\begin{figure}[htb]
	\centering
	\includegraphics[width=\linewidth]{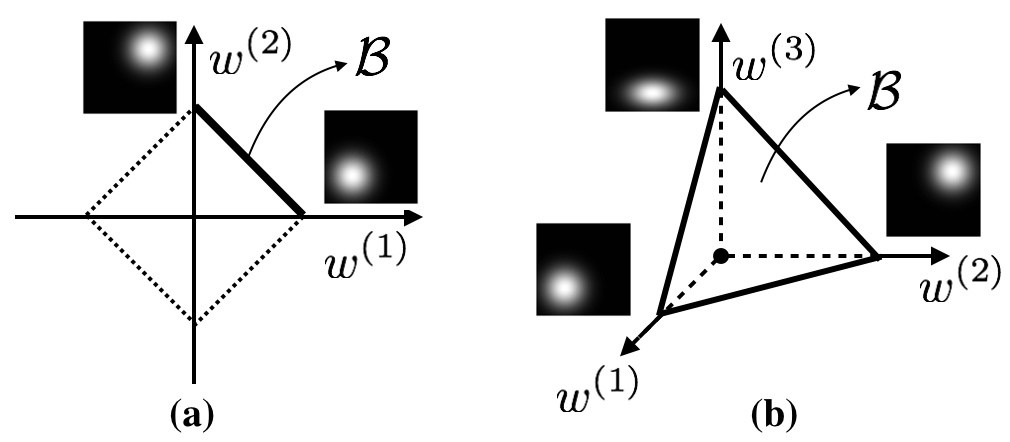}
	\caption{Examples of weight space $\mathcal{B}$ when (a) $m=2$ and (b) $m=3$. Symbol $w^{(i)}$ stands for the $i$-th component of a weight vector $\vec{w}$.}
	\label{moes:fig:weight_space}
\end{figure}

Let $\mathcal{B}:=\{\vec{w}, w^{(i)}>0,i=1,2,\dots,m, ||\vec{w}||_{1}=1\}$ denote the space of possible weight vectors (hereafter referred to as the \emph{weight space}), which is the first quadrant of the $m$-dimensional $\ell_1$-norm unit sphere. Examples of $\mathcal{B}$ when $m=2,3$ are shown in Fig.~\ref{moes:fig:weight_space}.
An info map $\phi$ can be decomposed with respect to (abbreviated as w.r.t.) a set of Fourier bases as $\phi=\Sigma_{k=0}^{K}\phi_k F_k$, where $\phi_k$ denotes the Fourier coefficient corresponding to each Fourier basis function $F_k, k=0,1,\dots,K$.
In practice, $K$ is often selected to be a finite number instead of infinity.
For a weight vector $\vec{w} \in \mathcal{B}$ and a set of info maps, the \emph{scalarized info map} can be represented as the weighted-sum of the corresponding Fourier coefficients:
\begin{align}
    \phi'&=\sum_{k=0}^{K}\phi'_k F_k \nonumber \\
    &=\sum_{i=1}^{m} w^{(i)}\phi^{(i)} = \sum_{i=1}^{m}w^{(i)} \left( \sum_{k=0}^{K}\phi^{(i)}_k F_k \right).
\end{align}
Then, for each $k=0,1,\dots,K$:
\begin{gather}
    \phi'_k = \sum_{i=1}^{m}w^{(i)}\phi^{(i)}_k = \vec{w}\cdot\Phi_k, \label{moes:eqn:phik_prime}
\end{gather}where $\Phi_k = (\phi^{(1)}_k,\phi^{(2)}_k,\dots,\phi^{(m)}_k)$, $\vec{w} = (w^{(1)},w^{(2)},\dots,w^{(m)})$, and $\cdot$ stands for the vector dot product.
The ergodic metric of a trajectory $q(t)$ (whose time averaged statistics is described as a set of Fourier coefficients $c_k$) w.r.t. $\phi'$ is
\begin{align}
    \mathcal{E}(\phi',q) & = \sum_{k=0}^{K}\lambda_k(c_k - \phi'_k)^2 \nonumber \\
    &=\sum_{k=0}^{K}\lambda_k(c_k - \vec{w}\cdot\Phi_k)^2\label{moes:eqn:erg_scala_map}
\end{align}

To obtain a set of trajectories whose ergodic vectors approximate the Pareto-optimal front, this work develops a framework (Fig.~\ref{moes:fig:algo_overview}): intuitively, in each planning episode, a $\vec{w}$ is sampled from $\mathcal{B}$ and a corresponding scalarized info map $\phi'$ is computed with Eqn~\ref{moes:eqn:phik_prime}.
Then, an ergodic trajectory w.r.t. $\phi'$ is planned by minimizing $\mathcal{E}(\phi',q)$ in Eqn.~\ref{moes:eqn:erg_scala_map}.
Note that, with a given initial state of the robot and a control $u(t)$, a unique trajectory can be specified (via the so-called forward simulation).
Thus, for presentation purposes, we use a control $u(t), t\in [0,T]$ to identify a trajectory, and let $u(t)|_{\vec{w}}$ denote the ergodic trajectory computed w.r.t. the scalarized info map based on $\vec{w}$.
In other words, for each $\vec{w} \in \mathcal{B}$, a corresponding ergodic trajectory  $u(t)|_{\vec{w}}$ can be computed.

Additionally, with Eqn.~\ref{moes:eqn:erg_scala_map}, we can observe that:
\begin{itemize}
    \item $\mathcal{E}(\phi',q)$ (i.e., the objective function to be minimized after scalarizing the info maps) is a \emph{convex} function w.r.t. $\vec{w}$ and $c_k$.
    \item Although $c_k$ is non-convex with respect to $u(t)$ due to the robot dynamics and the Fourier bases, existing ergodic search algorithms~\citep{mathew2011metrics,mavrommati2017real,miller2016ergodic} have shown that this non-convexity can be handled by iterative gradient descent optimization in practice.
\end{itemize}
Based on these observations, we take the view that a set of Pareto-optimal trajectories can be efficiently obtained by episodically sampling new $\vec{w}$ in the neighborhood of the current weight vector, and running local optimization in each episode.
Following this idea, we propose a framework called Sequential Local Ergodic Search (SLES), which is explained in the next section.

\subsection{Sequential Local Ergodic Search (SLES)}

Intuitively, SLES covers (or say explores) the weight space $\mathcal{B}$ from some initial weight vector $\vec{w}_{init}$ in a breadth-first manner in order to approximate the Pareto-optimal front.
SLES iteratively (i) scalarizes the info maps based on the current weight vector $\vec{w}$, (ii) leverages regular (single-objective) ergodic search to compute a trajectory (represented by $u(t)|_{\vec{w}}$), and (iii) samples new weight vectors $\vec{w}'$ from $\mathcal{B}$ in the neighborhood of $\vec{w}$ and use $u(t)|_{\vec{w}}$ as an initial guess to optimize the ergodic trajectory corresponding to $\vec{w}'$. The above process iterates until $\mathcal{B}$ has been ``fully covered'' by sampled weight vectors.

\begin{algorithm}[htbp]
	\caption{Pseudocode for SLES}\label{moes:alg:sles}
	\begin{algorithmic}[1]
		\State{$\vec{w}_{init} \gets$ \textit{InitWeight}()}
		\State{$u_{init}(t)|_{\vec{w}_{init}}= 0$}
		\State{OPEN $\gets \emptyset$, CLOSED $\gets \emptyset$, $\mathcal{S}\gets \emptyset$ }
		\State{Add $\vec{w}_{init}$ into OPEN}
		\While{OPEN is not empty}
		\State{$\vec{w} \gets$ OPEN.pop()}
		\State{Compute $\{\phi'_k, k=0,1\dots,K\}$ with $\vec{w}$ and Eqn.~\ref{moes:eqn:phik_prime}}
		\State{$u(t)|_{\vec{w}} \gets$ \textit{ErgodicSearch}($\{\phi'_k\}$, $u_{init}(t)|_{\vec{w}}$})
		\State{Add $\vec{w}$ into CLOSED}
		\State{Add $u(t)|_{\vec{w}}$ into $\mathcal{S}$}
		\ForAll{$\vec{w}'\in$ \textit{Neighbor}($\vec{w}$)}
		\If{$\vec{w}' \notin$ OPEN $\cup$ CLOSED}
		\State{$u_{init}(t)|_{\vec{w}'} \gets u(t)|_{\vec{w}}$ }
		\State{Add $\vec{w}'$ to OPEN}
		\EndIf
		\EndFor
		\EndWhile
		\State{\textbf{return} $\mathcal{S}$}
	\end{algorithmic}
\end{algorithm}

Specifically, as shown in Alg.~\ref{moes:alg:sles}, SLES begins by initializing a weight vector $\vec{w}_{init}$ (line 1), which can be either randomly sampled from $\mathcal{B}$, or specified by the user based on the domain knowledge of the specific application.
An initial control $u_{init}(t)|_{\vec{w}_{init}}$ corresponding to $\vec{w}$ is also initialized (line 2), which will later be used as the initial guess for the first episode of the ergodic search.
Let OPEN denote a first-in-first-out queue containing candidate weight vectors that need expansion, and expanding a weight vector $\vec{w}$ means computing $u(t)|_{\vec{w}}$ and sampling new weight vectors in the neighborhood of $\vec{w}$.
Let CLOSED denote a set of weight vectors that have been expanded, and let $\mathcal{S}$ denote the set of corresponding $u(t)|_{\vec{w}}$ for each $\vec{w} \in$ CLOSED that have been computed at any time during the computation.
Initially, OPEN, CLOSED and $\mathcal{S}$ are all initialized as empty sets (line 3) and then $\vec{w}_{init}$ is added to OPEN (line 4).

In each planning \emph{episode} (lines 5-14), a weight vector $\vec{w}$ is popped from OPEN and the corresponding scalarized info map $\phi'$ is computed based on Eqn.~\ref{moes:eqn:phik_prime}, which is represented by its Fourier coefficients (line 7).
Then, a regular ergodic search algorithm is invoked to cover $\phi'$, which iteratively minimizes Eqn.~\ref{moes:eqn:erg_scala_map} from the initial guess $u_{init}(t)|_{\vec{w}}$ (see Sec.~\ref{moes:sec:ergodic_search}).
The computed solution trajectory (represented by $u(t)|_{\vec{w}}$) as well as the corresponding $\vec{w}$ are then added to $\mathcal{S}$ and CLOSED respectively.
Finally, neighbor weight vectors of $\vec{w}$ in $\mathcal{B}$ are sampled (see Sec.~\ref{moes:sec:ngh_sample_basic} and \ref{moes:sec:ngh_sample_adaptive}) and is represented by \textit{Neighbor}($\vec{w}$).
For each $\vec{w}' \in$ \textit{Neighbor}($\vec{w}$), if $\vec{w}'$ has not been generated yet (i.e., $\vec{w}' \notin $ OPEN$\cup$CLOSED), $\vec{w}'$ is added to OPEN for future expansion.

SLES terminates when OPEN is empty, which indicates that $\mathcal{B}$ has been fully covered by the sampled weight vectors.
At termination, $\mathcal{S}$ is returned (line 15), which contains a set of controls, each of which specifies a trajectory, and the ergodic vectors of all these trajectories provide an approximation to the Pareto-optimal front.

\subsection{Ergodic Search Procedure}\label{moes:sec:ergodic_search}
A benefit of SLES is its ability to leverage existing ergodic search algorithms to cover $\phi'$ in procedure \textit{ErgodicSearch} in each planning episode.
This work leverages the existing approach in~\citep{abraham2018decentralized}, which iteratively minimizes the ergodic metric as introduced in Eqn.~\ref{moes:eqn:erg_scala_map} within an optimal control framework, and is able to handle general non-linear dynamics of the robot.
Other ergodic planners, such as~\citep{mathew2011metrics,mavrommati2017real}, can also be used to implement \textit{ErgodicSearch} within the framework of SLES.

In each planning episode, SLES invokes \textit{ErgodicSearch} with a specific initial guess $u_{init}|_{\vec{w}}$, instead of using a random or zero control as the initial guess.
Specifically, this initial guess is set to be a solution $u(t)|_{\vec{w}'}$ computed in the previous episodes, whose corresponding weight vector $\vec{w}'$ is close to the current weight vector $\vec{w}$ in the weight space $\mathcal{B}$.
As shown in Sec.~\ref{moes:sec:result}), this ``local optimization'' strategy expedites the overall planning in comparison with a naive scalarization method, which uses $u(t)=0$ as the initial guess for each episode.

\subsection{Basic Version of Neighbor Sampling}\label{moes:sec:ngh_sample_basic}
\begin{figure}[htb]
	\centering
	\includegraphics[width=0.8\linewidth]{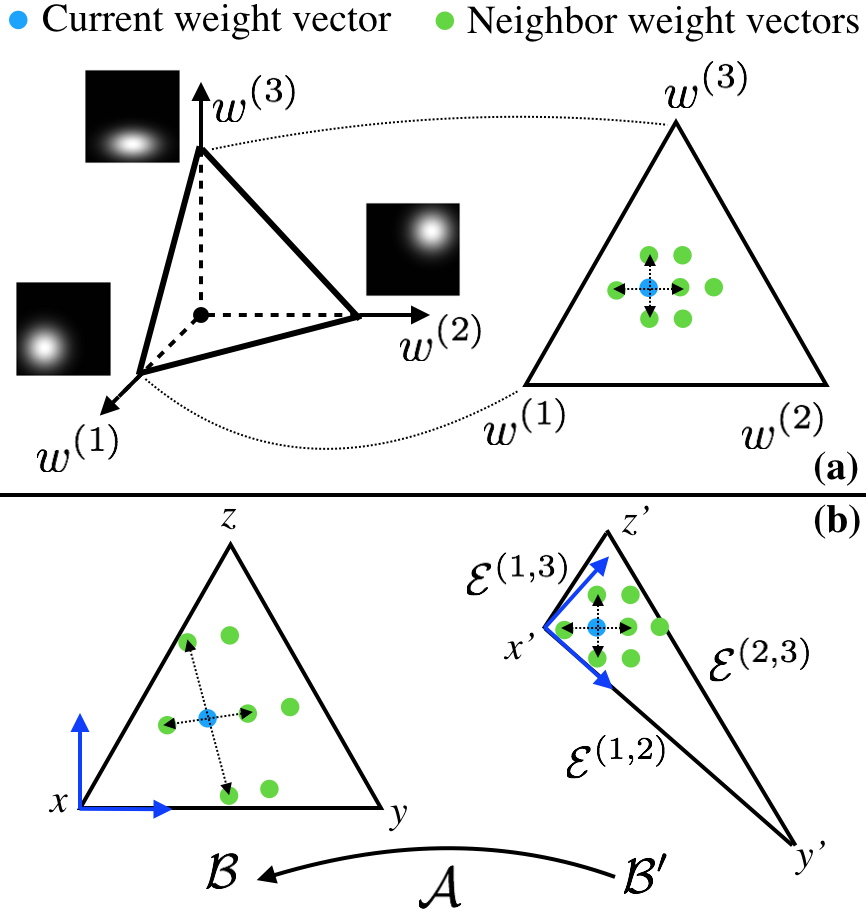}
	\caption{(a) shows the basic sampling method in the weight space $\mathcal{B}$. (b) shows the adaptive sampling method in the affine transformed weight space $\mathcal{B}'$. Each sampled point in $\mathcal{B}'$ can be affine transformed to a (valid) weight vector in $\mathcal{B}$.}
	\label{moes:fig:adaptive_sampling}
\end{figure}

While SLES is general to arbitrary $m > 1$, to simplify the presentation, we limit our focus to $m=2,3$.
Given a weight vector $\vec{w} \in \mathcal{B}$, this work takes a deterministic sampling strategy with a hyper-parameter $d$ denoting the sampling step size.
When $m=2$, $\mathcal{B}$ is a closed line segment, and the neighbors of a given $\vec{w}$ are defined to be the weight vectors that is of distance $d$ away from $\vec{w}$ along the line segment.
When $m=3$, $\mathcal{B} \subset \mathbb{R}^2$ is the closed set enclosed by a triangle as shown in Fig.~\ref{moes:fig:weight_space} (b).
The neighbors of a given $\vec{w}$ are defined to be the four weight vectors that are of distance $d$ away from $\vec{w}$ along the four cardinal directions, as shown in Fig.~\ref{moes:fig:adaptive_sampling} (a).

In general, $\mathcal{B}$ is the first quadrant of the $m$-dimensional $\ell_1$-norm unit sphere, which is a $(m-1)$-dimensional bounded closed set.
Using the geometry term, $\mathcal{B}$ is a $(m-1)$-simplex, and each corner point of $\mathcal{B}$ corresponds to an info map to be covered.
Since $\mathcal{B}$ is bounded, the aforementioned deterministic sampling strategy generates a finite number of weight vectors from $\mathcal{B}$, and SLES is guaranteed to terminate when all these sampled weight vectors are expanded.
Additionally, this sampling method can be generalized to $m > 3$.
However, the total number of possible samples grows exponentially w.r.t. $m$, and we leave this potential scalability issue (when $m$ is large) to our future work.

A limitation of this deterministic sampling strategy is that it does not consider the ``similarity'' between info maps to be covered.
For example, if two info maps to be covered  (i.e., two objectives) are similar (or very different) to each other, then only a few (or a lot of) weight vectors are needed to obtain a good approximation of the Pareto-optimal front.
We handle this limitation in the ensuing section.

\subsection{Adaptive Neighbor Sampling}\label{moes:sec:ngh_sample_adaptive}
This section develops an \emph{adaptive} neighbor sampling method, which can adjust the ``density'' of samples based on the similarity of info maps to be covered.
Let $\mathcal{E}^{(i,j)}$ denote the ergodic metric between two info maps $\phi^{(i)},\phi^{(j)}$:
\begin{gather}
    \mathcal{E}^{(i,j)} = \sum_{k=0}^{K} \lambda_k (\phi^{(i)}_k - \phi^{(j)}_k)^2,
\end{gather}
which characterizes the difference between $\phi^{(i)}$ and $\phi^{(j)}$ using Fourier coefficients.
For example, in Fig.~\ref{moes:fig:result_o3} (a), info maps $\phi^{(1)}$ and $\phi^{(3)}$ are similar to each other ($\mathcal{E}^{(1,3)}$ is small) while $\phi^{(1)}$ and $\phi^{(2)}$ are different from each other ($\mathcal{E}^{(1,2)}$ is large).

Then, an \emph{affine transformed weight space} $\mathcal{B}'$ is constructed as follows.
Let $\mathcal{B}'$ be an $(m-1)$-simplex where (i) each corner point of $\mathcal{B}'$ corresponds to an info map $\phi^{(i)}$, and (ii) the line segment connecting two corner points (corresponding to $\phi^{(i)}$ and $\phi^{(j)}$) has length $\mathcal{E}^{(i,j)}$.
$\mathcal{B}'$ exists as the ergodic metric is a Sobolev metric~\citep{mathew2011metrics} and satisfies the triangle inequality.\footnote{It is possible that $\mathcal{B}'$ degenerates and is of dimension less than $(m-1)$. (For example, when $m=3$, the $(m-1)$-simplex is a triangle. If the three corner points of the triangle are co-linear, the triangle degenerates into a line segment.) For a degenerate case, the proposed adaptive sampling is not applicable while the basic sampling in Sec.~\ref{moes:sec:ngh_sample_basic} still works. For the rest of the presentation, we consider the case where the constructed $(m-1)$-simplex is non-degenerate.}

After specifying a coordinate system to both $\mathcal{B}'$ and $\mathcal{B}$, an affine transformation $\mathcal{A}: \mathcal{B}' \rightarrow \mathcal{B}$ can be found by associating each pair of corner points $(p,p'), p\in\mathcal{B},p'\in\mathcal{B}'$.
For each $p'\in \mathcal{B}'$, a corresponding point $\mathcal{A}(p') \in \mathcal{B}$ can be found, and the corresponding weight vector $\vec{w}$ can be obtained based on the coordinate of $\mathcal{A}(p')$.
For presentation purposes, let $\mathcal{A}_{\vec{w}}(p'), p'\in \mathcal{B}'$ denote the map from the coordinate of a point $p' \in \mathcal{B}'$ to the actual weight vector $\vec{w}$ that will be used to scalarize the info maps, and let $\mathcal{A}_{\vec{w}}^{-1}(\vec{w}), \vec{w} \in \mathcal{B}$ denote the inverse map from a weight vector $\vec{w}$ to the coordinate of a point in $\mathcal{B}'$.

An example when $m=3$ is shown in Fig.~\ref{moes:fig:adaptive_sampling} (b). $\mathcal{B}'$ is a triangle, where the three enclosing line segments have lengths $\mathcal{E}^{(1,2)},\mathcal{E}^{(2,3)},\mathcal{E}^{(3,1)}$ respectively.
A possible coordinate system for $\mathcal{B}'$ is to place the origin at point $x' \in \mathcal{B}'$, align the {x}-axis with line segment $x'y'$. Then point $y'$ has coordinate $(\mathcal{E}^{(1,2)},0)$ and the coordinate of point $z'$ can be determined since the length of $x'z'$ and $y'z'$ are both known.
With equations $x=\mathcal{A}(x'),y=\mathcal{A}(y'),z=\mathcal{A}(z')$, the affine map $\mathcal{A}$ can be determined.

\begin{algorithm}[htbp]
	\caption{Pseudocode for \textit{AdaptiveNeighbor}($\vec{w}$)}\label{moes:alg:ngh}
	\begin{algorithmic}[1]
		\State{$O \gets \emptyset$}\Comment{The output, a set of weight vectors.}
		\State{$p' \gets \mathcal{A}_{\vec{w}}^{-1}(\vec{w})$}
		\State{$\Delta \gets \{(0,d'),(0,-d'),(d',0),(-d',0)\}$}\Comment{$m=3$}
		\ForAll{$\delta \in \Delta$}
		\State{$p'_{new} \gets p' + \delta$}
		\If{$p'_{new} \notin \mathcal{B}'$}
		\State{\textbf{continue}}
		\EndIf
		\State{Add $\mathcal{A}_{\vec{w}}(p'_{new})$ to $O$}
		\EndFor
		\State{\textbf{return} $O$}
	\end{algorithmic}
\end{algorithm}

With $\mathcal{B}'$ and the map $\mathcal{A}_{\vec{w}}$, SLES can sample points from $\mathcal{B}'$ (instead of directly sampling weight vectors from $\mathcal{B}$), and each sampled point $p'\in \mathcal{B}'$ can be transformed into a (valid) weight vector $\vec{w} \in \mathcal{B}$, which is then used to compute a scalarized info map.
Specifically, as shown in Alg.~\ref{moes:alg:ngh}, let $O$ denote the set of sampled weight vectors, which is initialized as an empty set, and let $\Delta$ denote the set of possible differences between two neighboring points in $\mathcal{B}'$ using the aforementioned deterministic sampling strategy.
Here, we use $d'$ to denote the step size to note the difference from the $d$ in the previous section.
(Note that, line 3 in Alg.~\ref{moes:alg:ngh} only shows the $\Delta$ when $m=3$.)
For each $\delta\in\Delta$, a neighbor point $p'_{new}\gets p' + \delta$ is generated.
If $p'$ is still within $\mathcal{B}'$, the corresponding weight vector $\mathcal{A}_{\vec{w}}(p'_{new})$ is added to $O$.
Finally, set $O$ is returned, which contains all sampled neighbor weight vectors.
Here, $\mathcal{B}'$ is constructed by considering the difference between info maps, and sampling from $\mathcal{B}'$ allows SLES to \emph{adapt} the sampling density to the difference between info maps, which is verified in Sec.~\ref{moes:sec:result}.

\subsection{Discussion}\label{moes:sec:discussion}
\subsubsection{Earlier Termination}
In practice, when a strong prior preference between info maps (represented by $\vec{w}_{init}$) is available, the termination condition of SLES can be modified so that SLES terminates earlier when a certain neighborhood of $\vec{w}_{init}$ has been explored.
Note that SLES explores the weight space $\mathcal{B}$ in a breadth-first manner, and thus SLES explores the neighborhood around $\vec{w}_{init}$ in $\mathcal{B}$ at first.
This allows SLES to quickly compute an approximated Pareto-optimal front that is ``centered'' on the prior preference of the user.

\subsubsection{Weight Space Coverage}
SLES can be regarded as a framework that converts a MOES problem into a ``weight space coverage problem'': SLES iteratively samples weight vectors from $\mathcal{B}$ and terminates when the entire weight space $\mathcal{B}$ is covered.
The adaptive sampling method transforms $\mathcal{B}$ into $\mathcal{B}'$ based on the ergodic metrics between info maps and then covers $\mathcal{B}'$.
Note that other types of transformation (e.g. non-linear transformation) or different sampling methods may also be leveraged based on the domain knowledge of the application within the proposed SLES framework.

\section{Numerical Results}\label{moes:sec:result}

\subsection{Baseline Methods and Implementation}

MOGAs \citep{deb2002fast,emmerich2018tutorial} are popular approaches to solve MOO problems, which are also applicable to the MOES problem in this work.
For comparison, this work uses NSGA-II~\citep{deb2002fast}, a popular MOGA for MOO problems, as the first baseline approach.\footnote{NSGA-II is popular for MOO problems with two or three objectives. When there are more than three objectives (sometimes referred to as ``many-objective optimization''), NSGA-III can be used.}
A second baseline approach considered in this work is a naive scalarization method, which differs from SLES as it leverages neither the idea of sequential local optimization nor adaptive weight sampling.
It iteratively samples $\vec{w}\in \mathcal{B}$, and plans ergodic trajectory w.r.t. the scalarized info map by optimizing from some common naive initial guess, such as a zero control input.

We implement our algorithms\footnote{Our code is available at \url{https://github.com/wonderren/public_moes}} and the naive scalarization method in Python, and use the NSGA-II implementation from pymoo~\citep{pymoo}, a MOGA library, for our experiments.
We run tests on a laptop with an Intel Core i7 CPU and 16 GB RAM.
All tests have a workspace of size $[0,1]\times[0,1]$.
We specify the robot dynamics as a forward-moving-only differential-drive robot that initially locates at the center of the workspace $(0.5, 0.5)$ with orientation zero (pointing to the right).
For presentation purposes, we use ``Scala.'' to denote the naive scalarization method, ``SLES'' to denote our algorithm with the basic neighbor sampling method (Sec.~\ref{moes:sec:ngh_sample_basic}), and ``A-SLES'' to denote our algorithm with the adaptive neighbor sampling method (Sec.~\ref{moes:sec:ngh_sample_adaptive}).
We set a termination threshold $\epsilon =10^{-3}$ for each \textit{ErgodicSearch} call in Alg.~\ref{moes:alg:sles}: when the ergodic metric w.r.t. the scalarized info map (Eqn.~\ref{moes:eqn:erg_scala_map}) is no larger than $\epsilon$, \textit{ErgodicSearch} terminates.

To describe the approximation quality of the Pareto-optimal front, we use the ``hyper-volume'' indicator (H.V.)~\citep{emmerich2018tutorial} from the MOO literature.
Intuitively, H.V. denotes the volume enclosed by the approximated Pareto-optimal front and a reference point in the objective space, which is set to $(1,1,\dots,1)$ in this work.

\subsection{Comparison with NSGA-II}
\begin{figure}[t!]
	\centering
	\includegraphics[width=\linewidth]{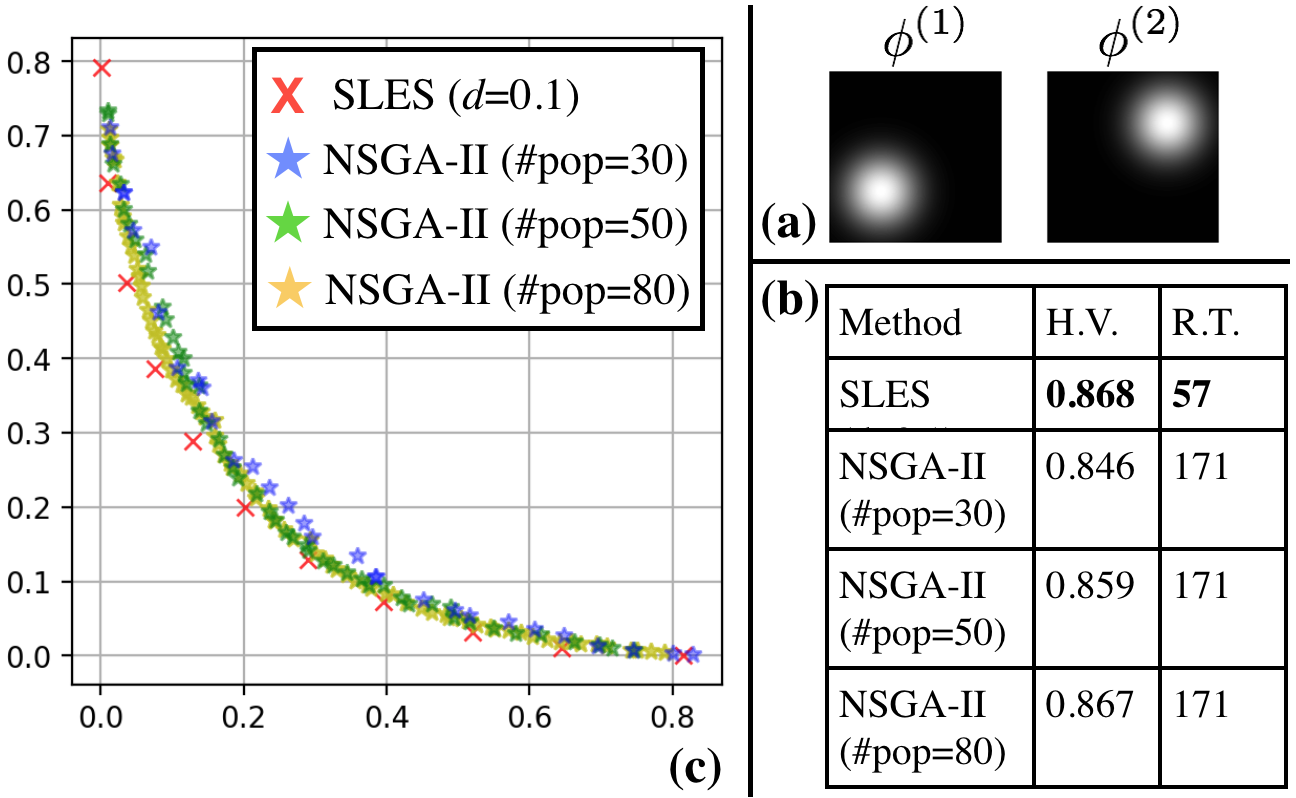}
	\caption{(a) shows the two info maps to be covered. (b) shows the hyper-volume (H.V.) of the solution set computed by each method, where we allow NSGA-II (baseline) to run for \emph{three} times the run time (R.T., in seconds) of SLES. (c) visualizes the ergodic vectors of the computed solutions. SLES computes a set of solutions with similar or better quality than NSGA-II while using only one third of the run time of NSGA-II.}
	\label{moes:fig:result_nsga2}
\end{figure}

\begin{figure*}
	\centering
	\includegraphics[width=\linewidth]{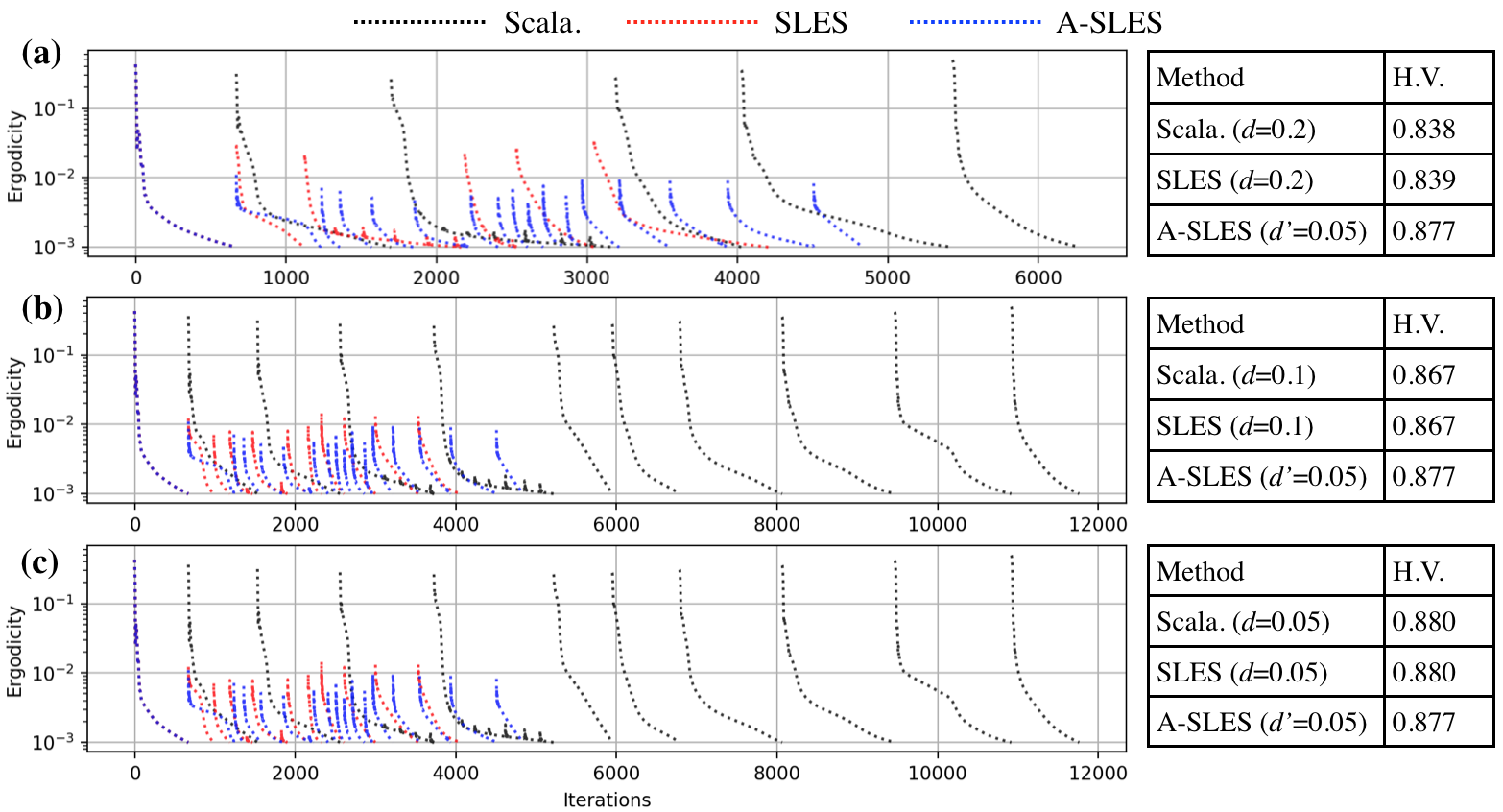}
	\caption{The horizontal axis indicates the number of optimization iterations in the \textit{ErgodicSearch} procedure while the vertical axis denotes the ergodic metric in  Eqn.~\ref{moes:eqn:erg_scala_map}. Note that at the beginning of each episode, a different $\vec{w}$ (and thus a different $\phi'$) is considered, and thus the ergodic metric ``jumps''. The corresponding tables in (a), (b) and (c) show the hyper-volumes of different methods with different step sizes.
	This figure shows that SLES requires obviously less computational time than Scala. (i.e., baseline) to compute a set of solutions with similar quality in terms of H.V. More discussion can be found in the text.}
	\label{moes:fig:result_sles_iters}
\end{figure*}

We begin our tests with $m=2$, and the info maps are shown in Fig.~\ref{moes:fig:result_nsga2} (a).
We compare SLES with NSGA-II. We measure the run time of SLES (denoted as $T_1$) and let NSGA-II run for \emph{three} times the run time of SLES (i.e., $3T_1$).
As shown in Fig.~\ref{moes:fig:result_nsga2}, increasing the ``population size'' (a hyper-parameter in NSGA-II) can slightly improve the solution quality.
However, SLES computes a set of solutions with similar or better quality (in terms of H.V.) than NSGA-II while using only one third of the run time of NSGA-II.
The possible reason is, while being general to various problems, NSGA-II treats the objective functions as a ``black-box'' and often ignores the underlying structure of the problem (such as the dynamics of the robot and the local metric structures).

\subsection{Comparison with Naive Scalarization}

We then compare SLES against the naive scalarization method (Scala.) with the same test settings as in the previous section.
In Fig.~\ref{moes:fig:result_sles_iters}, the horizontal axis indicates the number of optimization iterations in the \textit{ErgodicSearch} procedure while the vertical axis denotes the ergodic metric in  Eqn.~\ref{moes:eqn:erg_scala_map}.
Note that at the beginning of each episode, a different $\vec{w}$ (and thus a different $\phi'$) is considered, and thus the ergodic metric changes.
As shown in Fig.~\ref{moes:fig:result_sles_iters}, Scala. takes the most number of optimization iterations in each episode since it always starts from the same naive initial guess (i.e., a zero control input).
Both SLES and A-SLES run faster than Scala. especially when $d$ decreases (which means there are more planning episodes).
Take Fig.~\ref{moes:fig:result_sles_iters} (b) for example, SLES requires less than half of the run time in comparison with Scala., and still computes a solution set with the same quality in terms of H.V.
It shows that running local optimization by (i) sampling weight vectors that are near to each other, and (ii) reusing the solution from the previous episodes as the initial guess for the current episode, can expedite the computation.

Fig.~\ref{moes:fig:result_sles_iters} also demonstrates the benefit of the proposed adaptive neighbor sampling: specifying $d$ in $\mathcal{B}$ is not intuitive and can lead to either too sparse ($d=0.2$) or too dense sampling ($d=0.05$), which leads to either a low H.V. value or a large number of episodes.
Sampling based on $d'$ in the affine transformed weight space $\mathcal{B}'$ allows the algorithm to adapt to the differences between info maps.
Additionally, $d'$ has the same unit as the ergodic metric between info maps, and is thus more intuitive to specify.

\subsection{Different Sampling Step Sizes}
\begin{figure}[htb]
	\centering
	\includegraphics[width=0.9\linewidth]{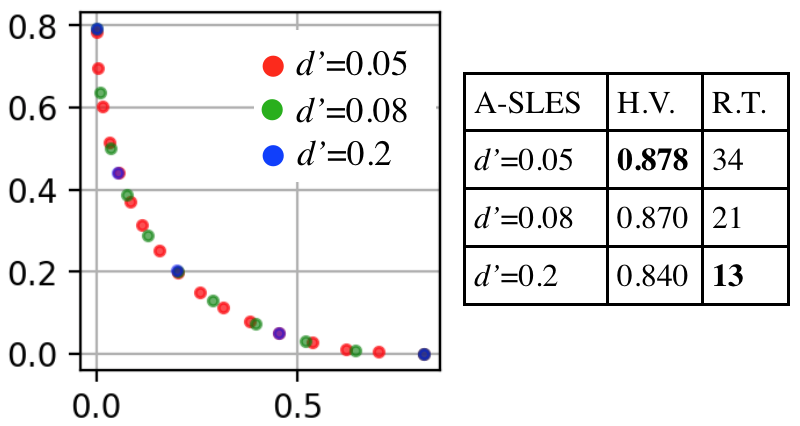}
	\caption{Hyper-volume (H.V.) and run time (R.T.) (in seconds) of A-SLES with varying sampling step size $d'$. This figure shows, by tuning $d'$, A-SLES can trade-off between solution quality and run time. Having slightly larger $d'$ can speed up the computation with small decrease in the H.V.}
	\label{moes:fig:result_asles_d}
\end{figure}

This section tests A-SLES with varying step sizes $d'$, with $m=2$, and with the same info maps as in the previous section.
As shown in Fig.~\ref{moes:fig:result_asles_d}, by tuning $d'$, there is a trade-off between H.V. values, which indicate the quality of the approximation, and the computational burden, which is indicated by the run time. Having slightly larger $d'$ can speed up the computation significantly with small decrease in H.V.

\subsection{Three Objectives}

\begin{figure}[htb]
	\centering
	\includegraphics[width=\linewidth]{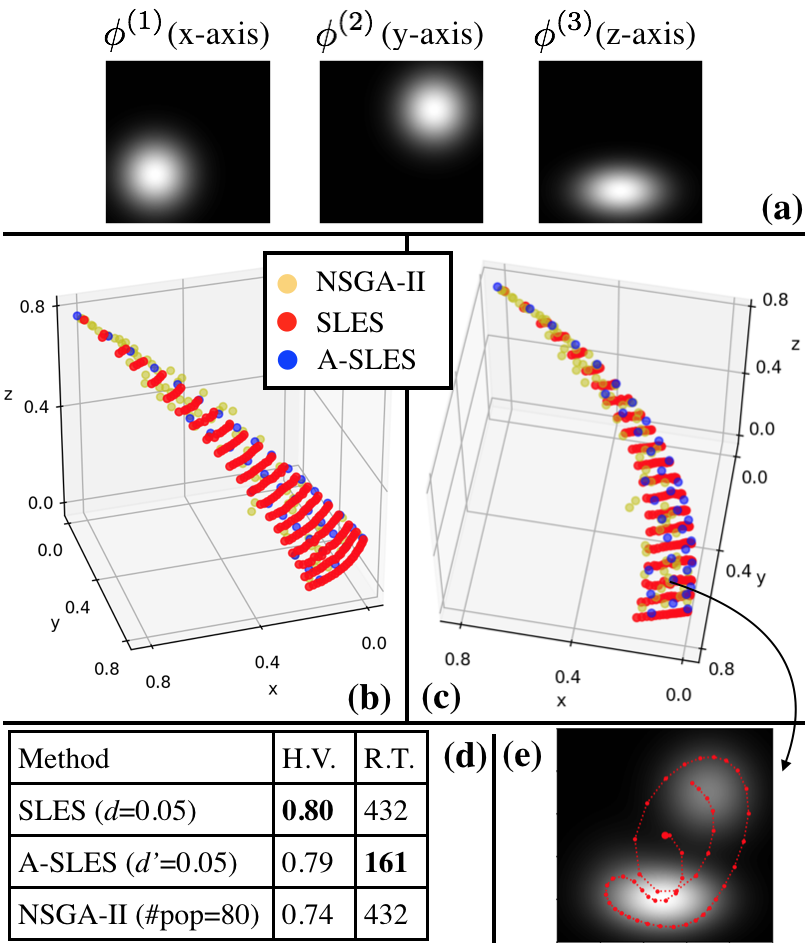}
	\caption{
	(a) shows the three info maps to be covered.
	(b) highlights the solution (red) computed by SLES and (c) highlights the solution (blue) computed by A-SLES.
	(d) shows the hyper-volume and run time (in seconds) of each method.
	(e) shows a scalarized info map and the corresponding ergodic trajectory.
	A-SLES computes solutions of similar quality while requiring less than half of the run time in comparison with SLES and NSGA-II.
	}
	\label{moes:fig:result_o3}
\end{figure}

Finally, we test NSGA-II, SLES and A-SLES with $m=3$. The info maps are shown in Fig.~\ref{moes:fig:result_o3} (a).
Note that $\phi^{(1)}$ is similar to $\phi^{(3)}$ while they are both quite different from $\phi^{(2)}$.
Fig.~\ref{moes:fig:result_o3} (d) shows that A-SLES provides an approximation of similar quality in comparison with SLES (in terms of H.V.) while having a much smaller run time.
From Fig.~\ref{moes:fig:result_o3} (b) and (c), it is obvious that A-SLES can adaptively sample weight vectors based on the difference between each pair of info maps: there are only a few blue points in Fig.~\ref{moes:fig:result_o3} (c) to approximate the Pareto-optimal front.
In contrast, SLES has a lot of samples (the red points in Fig.~\ref{moes:fig:result_o3} (b)) to approximate the Pareto-optimal front.

\subsection{Robot Simulation}

\begin{figure}[htb]
	\centering
	\includegraphics[width=\linewidth]{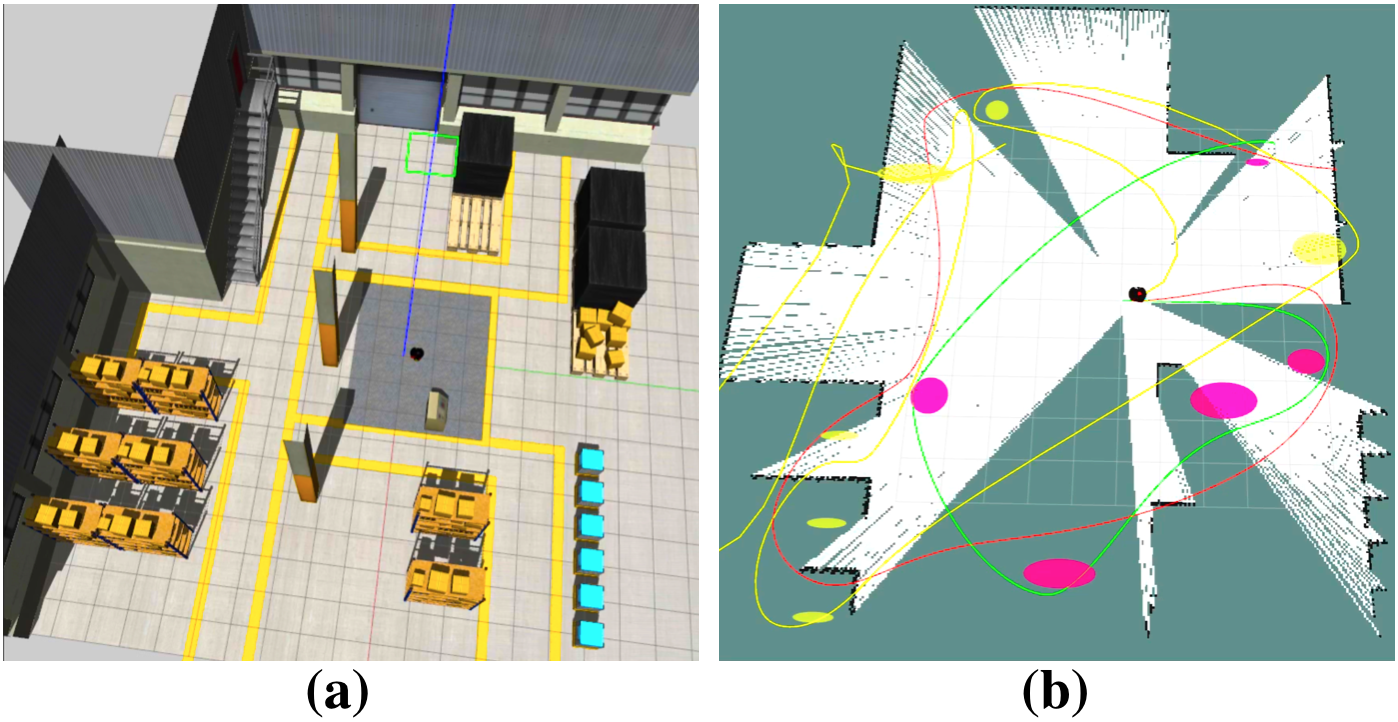}
	\caption{(a) shows the warehouse environment and (b) shows the information maps visualized as the yellow (probable gas leakage locations) and pink (probably survivor locations) markers on RViz. The current method does not consider obstacle avoidance during the ergodic planning and our simulation relies on an additional local planner to avoid obstacles.}
	\label{moes:fig:result_ros_sim}
\end{figure}

We apply the proposed A-SLES algorithm to an example MOES problem and simulate the computed trajectory in ROS.\footnote{Our ROS implementation leverages \url{https://github.com/wh200720041/warehouse_simulation_toolkit} and \url{https://github.com/bostoncleek/ergodic_exploration}.}
The example involves a search and rescue mission in a warehouse with hazardous gas leakage.
The goal is to find both sources of leakage and search for survivors.
The two objectives are described using two info maps, which can be generated based on the prior knowledge of the warehouse (Fig.~\ref{moes:fig:result_ros_sim}).
Usually these two objectives cannot be optimized simultaneously as survivors can be far away from the gas leakage source.
We use A-SLES to compute a set of Pareto-optimal trajectories, which can then be visualized to the decision maker on site so that a more informed decision can be made.
For example, if the effect of the gas for humans is minor but it affects the goods in the warehouse significantly, one might want to choose a trajectory that prioritizes finding the leakage source more than searching for humans inside.

Fig.~\ref{moes:fig:result_ros_sim} (b) visualizes three Pareto-optimal solutions. For instance, the green trajectory prioritizes finding survivors (the pink info map) while the red one favors localizing leakage sources (the yellow info map). Please refer to our multi-media attachment (\url{https://youtu.be/SEkwti-pGjE}) for more details.

\section{Conclusion and Future Work}\label{moes:sec:conclusion}
This work formulates a Multi-Objective Ergodic Search (MOES) problem, which requires planning trajectories to simultaneously cover multiple information maps.
To solve the MOES problem, a framework called Sequential Local Ergodic Search (SLES) is proposed, which explores the weight space (the space that contains all possible weight vectors) in a breadth-first manner by (i) sampling new weight vectors in the neighborhood of the current weight vector, and (ii) optimize the trajectory corresponding to the new weight vector by using the current solution as the initial guess.
Additionally, to further expedite SLES, we also develop a variant called Adaptive SLES (A-SLES) that can adjust the density of sampled weight vectors based on the ergodic metric between each pair of info maps to be covered. 
The numerical results verify the advantages of SLES and A-SLES over baselines.

This work is a first step to investigate MOES problems.
The current work considers one robot with multiple static info maps without obstacles.
It's worthwhile to investigate the MOES problems where the info maps are dynamic (i.e., updated in an online manner) or there are multiple robots.
Additionally, one can also consider using different ergodic search algorithms~\citep{ayvali2017ergodic,mavrommati2017real} within the framework of SLES to explicitly handle obstacle avoidance constraints or to achieve real-time planning.

\bibliographystyle{plainnat}
\bibliography{references}

\end{document}